\documentclass[runningheads,a4paper]{llncs}

\usepackage{wrapfig}

\usepackage{graphicx}               
\usepackage[intlimits]{amsmath}     
\usepackage{amssymb}                
\usepackage{amsfonts}               
\usepackage{enumitem}               
\usepackage{siunitx}                
\usepackage{url}                    
\usepackage{xspace}                 
\usepackage[T1]{fontenc}            
\usepackage[colorlinks=true,linkcolor=black,citecolor=black,urlcolor=black]{hyperref}    
\usepackage{textcomp}               
\usepackage{algorithm, algorithmic} 
\usepackage{array}                  
\usepackage{eqparbox}               

\graphicspath{{images/}}
\DeclareGraphicsExtensions{.pdf,.jpeg,.png,.jpg}

\providecommand{\norm}[1]{\lVert#1\rVert}

\newcommand{\degreem}{^{\circ}} 

\newcommand{\seclabel}[1]{\label{sec:#1}}

\newcommand{\figlabel}[1]{\label{fig:#1}}

\newcommand{\tablabel}[1]{\label{tab:#1}}
\newcommand{\eqnlabel}[1]{\label{eqn:#1}}

\newcommand{\figref}[1]{Fig.~\ref{fig:#1}\xspace}
\newcommand{\tabref}[1]{Table~\ref{tab:#1}\xspace}

\newcommand{\eqnref}[1]{(\ref{eqn:#1})\xspace}

\newcommand{\iguhop}{igus\textsuperscript{\tiny\circledR}$\!$ Humanoid Open Platform\xspace}

\newcommand{\degree}{$\degreem$\xspace}

\usepackage{eso-pic}

\AtBeginDocument{\AddToShipoutPictureFG*{\AtTextUpperLeft{\put(0,\LenToUnit{40pt}){\parbox{\textwidth}{\centering\bfseries
20th RoboCup International Symposium, Leipzig, Germany, 2016
}}}}}

\begin{document}

\mainmatter  

\title{Real-Time Visual Tracking and Identification for a Team of Homogeneous Humanoid Robots}

\titlerunning{Real-Time Visual Tracking and Identification}

\author{Hafez Farazi and Sven Behnke}
\authorrunning{H. Farazi and S. Behnke}

\institute{%
Autonomous Intelligent Systems, Computer Science Institute VI\\ University of Bonn, Germany\\
\path|{farazi, behnke}@ais.uni-bonn.de|\\
\url{http://ais.uni-bonn.de}}


\toctitle{Real-Time Visual Tracking and Identification} 

\maketitle

\begin{abstract}
The use of a team of humanoid robots to collaborate in completing a task is an 
increasingly important field of research. One of the challenges in achieving collaboration, 
is mutual identification and tracking of the robots. This work presents a 
real-time vision-based approach to the detection and tracking of robots of known 
appearance, based on the images captured by a stationary robot. A Histogram of Oriented Gradients 
descriptor is used to detect the robots and the robot 
headings are estimated by a multiclass classifier. The tracked robots report their own heading estimate from magnetometer readings. For 
tracking, a cost function based on position and heading is applied to each 
of the tracklets, and a globally optimal labeling of the detected robots is 
found using the Hungarian algorithm. The complete identification and tracking 
system was tested using two \iguhop robots on a soccer field. We expect that a 
similar system can be used with other humanoid robots, such as Nao and 
\mbox{DARwIn-OP}.
\end{abstract}

\section{Introduction}
\seclabel{introduction}

Multi-target tracking is a well-known problem in computer vision, and has many 
applications, including traffic monitoring and automated surveillance. The aim of 
multi-target tracking is to automatically find objects of interest, assign a 
unique identification number to each, and to follow their movements over time. 
Multi-target tracking is fundamentally different to single-target tracking 
because of the difference in the state space model used for each. In particular, 
data association in situations of multiple detections with closely spaced and/or 
occluded objects makes multi-target tracking significantly more difficult. The 
expected number of visible targets is often unknown and may vary over time.

This work addresses a problem with an additional level of 
difficulty---the identification and tracking of multiple robots of identical 
appearance. Despite the lack of visual clues, our system is not only able to 
track each detected robot, but also to identify which robots are being 
tracked. This is done by generating a cost function for each tracklet, based on 
a motion model and the differences between the set of estimated and broadcasted 
headings of the robots. The output of the system, which is an estimation of 
the location and heading of each robot, is made available to the robots 
being observed, so that they may incorporate this into their own 
localization estimates, or use it for the generation of cooperative behaviors.
\figref{overview} gives an overview of our system.

\noindent The main contributions of this paper include\vspace*{-1ex}:
\begin{enumerate}
\item The introduction of a novel pipeline to identify a set of homogeneous 
humanoid robots in an image.
\item The development of a high accuracy and low training time humanoid robot 
detection algorithm, based on a Histogram of Oriented Gradients descriptor.
\item A robust method for the estimation of the relative heading of a robot.
\item Experimental evidence that the proposed method can cope with long-term 
occlusions, despite a lack of visual differences between the tracked targets.
\item Demonstration that it is possible to track, identify and localize a 
homogeneous team of humanoid robots in real-time from another humanoid robot.
\end{enumerate}

\begin{figure}[!tb]
\vspace*{-1ex}
\centering
\hspace*{-2.65cm}
\includegraphics[width=17cm]{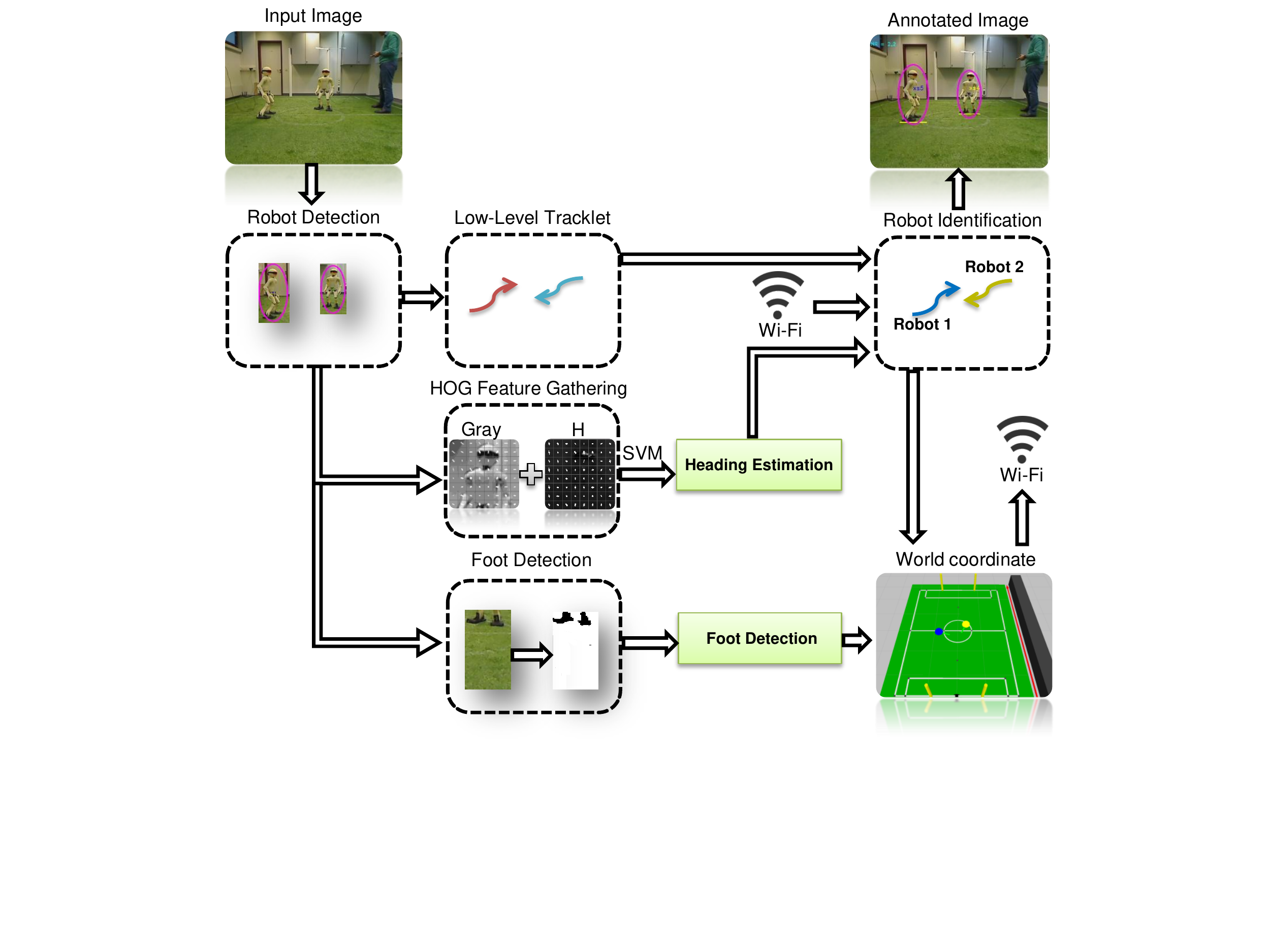}
\vspace*{-23ex}
\caption{Overview of our approach. After detection part, the heading of each robot is estimated based on proper HOG features. Using heading estimation and low-level tracklets observer finds and broadcasts the position of each robot.}
\figlabel{overview}
\vspace*{-2.8ex}
\end{figure}

\section{Related Work} 
\seclabel{related_work}
\vspace*{-1ex}

The related work is divided into three 
categories, \emph{multi-target tracking}, \emph{robot detection and tracking} and \emph{visual orientation 
estimation}. 

\textbf{Multi-target tracking} has been studied for many years in the field of 
computer vision. The tracking of targets in the absence of any 
category information is referred to as category free tracking (CFT) 
\cite{zhang2012robust}. CFT 
approaches normally do not require a detector that is trained offline, but rely on manual initialization.
Objects are tracked mainly based 
on visual appearance, and the system attempts to track each target by 
discriminating it from other regions of the image. The visual target model is 
usually updated online to cope with viewpoint and illumination changes. Two 
successful examples of the CFT approach include the works of Allen et al. 
\cite{allen2004object} and Yang et al. \cite{yang2005fast}. Although CFT 
approaches are computationally inexpensive and easy to implement, they are prone 
to excessive drift, after which it is very hard to recover.

Tracking by detection is one of the most popular approaches to multi-target 
tracking problems, as objects are naturally reinitialized when they are lost, 
and extreme model drifts cannot occur. As such, association based 
tracking (ABT) methods, which associate object detections with observed tracks, 
are proposed e.g. by Xing et al. \cite{xing2009multi}. An offline training procedure is generally used for the 
detection of objects of interest in each frame, and continuous object detections 
over time are linked to form so-called tracklets. Tracklets can then be 
associated with each other to form longer tracks. In most works, the probability 
of two tracklets being associated with each other is calculated based on a 
motion model and other criteria of visual similarity. The global tracklet 
association of highest probability is then computed using either the Hungarian 
algorithm \cite{xing2009multi}, a Markov chain Monte 
Carlo method \cite{yu2007multiple}, or a Conditional Random 
Field \cite{milan2015joint}.

\textbf{Robot detection and tracking} was done by Marchant et al. \cite{marchant2013cooperative} using both 
visual perception and sonar data which was targeted for soccer environment. However, 
anthropomorphic design requirements in the Humanoid League prohibit teams from using 
sonar sensors. 
Many object detection approaches cannot be used for robot 
detection tasks due to the limitation in the onboard computer. Arenas et al. \cite{arenas2007detection} detected Aibo robot and humanoid robots using the cascade of 
boosted classifiers, which is suitable for real-time applications. In another work Ruiz-Del-Solar et al. \cite{ruiz2010visual} proposed nested cascades of boosted 
classifiers for detecting legged robots. In addition to robot detection, 
gaze-direction of the robot is estimated based on Scale 
Invariant Feature Transform (SIFT) descriptor by Ruiz-Del-Solar et al. \cite{ruiz2010play}.

\textbf{Visual orientation estimation} of an object is often done by comparing 
projections of an accurate 3D model of the object to what is observed in the 
image, and finding the orientation that best matches the detected features 
\cite{dementhon1995model}. These approaches work 
best only on simple backgrounds. Since the background in our application can be 
quite cluttered, and we do not wish to rely on the existence of an accurate 3D 
model of the detected robot, the most suitable approach for orientation 
estimation is through the use of image descriptors. Lin et al. 
\cite{lin2009object} proposed an orientation recognition system based on a SIFT descriptor \cite{lowe1999object}, and a 
Support Vector Machine (SVM) classifier \cite{cortes1995support}. Shaikh et al. 
\cite{shaikh2012recognition} proposed a template-based orientation estimation method for 
images of cars, based on the comparison of shape signatures.

\section{Multi-Target Tracking Formulation}
\seclabel{multi-target_tracking_formulation}

We assume to have a collection of $N$ humanoid robots of identical 
appearance that need to be identified and tracked by a further standing robot, 
or a stationary camera. In each camera frame, each of the robots can either 
be fully visible, partially visible, or not visible at all, and may be 
performing soccer actions such as walking, kicking and getting up. As such, the 
durations of partial or total lack of visibility may either be short or long. 
Each robot is equipped with a 9-axis inertial measurement unit (IMU), and the 
estimated absolute heading of the robot is broadcasted over Wi-Fi. The Wi-Fi 
communication between the robots is assumed to have delays, data loss, and even 
potentially connection loss for up to a few seconds. We use NimbRo network library \cite{nimbro}
for Wi-Fi communication. Our objective is 
to detect, track and identify the $N$ robots based solely on the captured images 
and the broadcasted heading information. Two \iguhop robots were used for the
verification of the approach in this paper.

\section{Vision System}
\seclabel{vision_system}

\subsection{Robot Detection}
\seclabel{robot_detection}

Although a number of pre-trained person detectors are available online, there is 
no detector for humanoid robots that can work out of the box. As such, we have 
designed, implemented and tested a robot detector that can robustly detect the 
\iguhop, although we expect the detector to work for other humanoid robot 
model as well, with the appropriate retuning and retraining. We evaluated five different methods for their suitability in our target domain before selecting and refining 
the most promising one. The methods were based, respectively, on color 
segmentation \cite{Farazi2015}, adaptive object labeling \cite{wang2015online}, 
Haar wavelets \cite{haar}, Local Binary Patterns \cite{liao2007learning}, and 
Histograms of Oriented Gradients (HOG) \cite{zhu2006fast}. The last of the five, 
a HOG feature descriptor used in the form of a cascade classifier was chosen 
based on criteria such as detection rate and training time. Although many 
RoboCup teams use a simple color segmentation approach to robot and obstacle 
detection, this approach is not safe in our case because we want to be able to 
distinguish the \iguhop from other objects on the field, such as the referee. 
Adaptive object labeling produced a relatively high rate of false positives, and 
it was nearly impossible to find a suitable threshold to work at all distances and 
in all situations. The method was also not able to deal well with occlusions. 
The Local Binary Pattern-based feature classifier also produced relatively poor 
results. However, the overall detection rates for the Haar wavelet and HOG 
cascading methods were found to be relatively good, and quite similar, but the 
former required a significantly longer time to train, so the latter was chosen.

In contrast to what is suggested for pedestrian detection 
\cite{dalal2006object}, we do not feed the output of a multi-scale sliding 
window to a support vector machine (SVM) classifier. Instead, to save on 
computation time we use a cascade of rejectors with the AdaBoost technique to 
choose which features to evaluate in each stage, similar to what is suggested by Zhu et al. 
\cite{zhu2006fast}. By using HOG features, we obtain a description of the visual 
appearance of the robot that is invariant to changes in illumination, position, 
orientation and background. As HOG is not rotation-scale invariant however, we 
artificially expand the number of positive images used for training by applying 
a number of transformations, also in part to minimize the required user effort 
in gathering the samples. These transformations include random rotations up to 
\textpm\SI{15}{\degree}, mirroring, and the cutting of some parts of the sample 
image, in particular at the bottom, left and right, to emulate partial 
occlusion. Note that larger rotations of the images are not applied to allow the 
classifier to learn the shadow under the robot. This also has the positive 
effect of not detecting sitting or fallen robots, so that this discrete 
difference can be used in the identification phase to discern the robots. In our training of the 
\iguhop, we used a set of about 500 positive samples, 1000 negative samples, 
and a cascade classifier with 20 stages. The training time for the 
classifier was about \SI{12}{\hour} on a standard PC.

\begin{figure}[!tb]
\centering
\includegraphics[width=2.35cm]{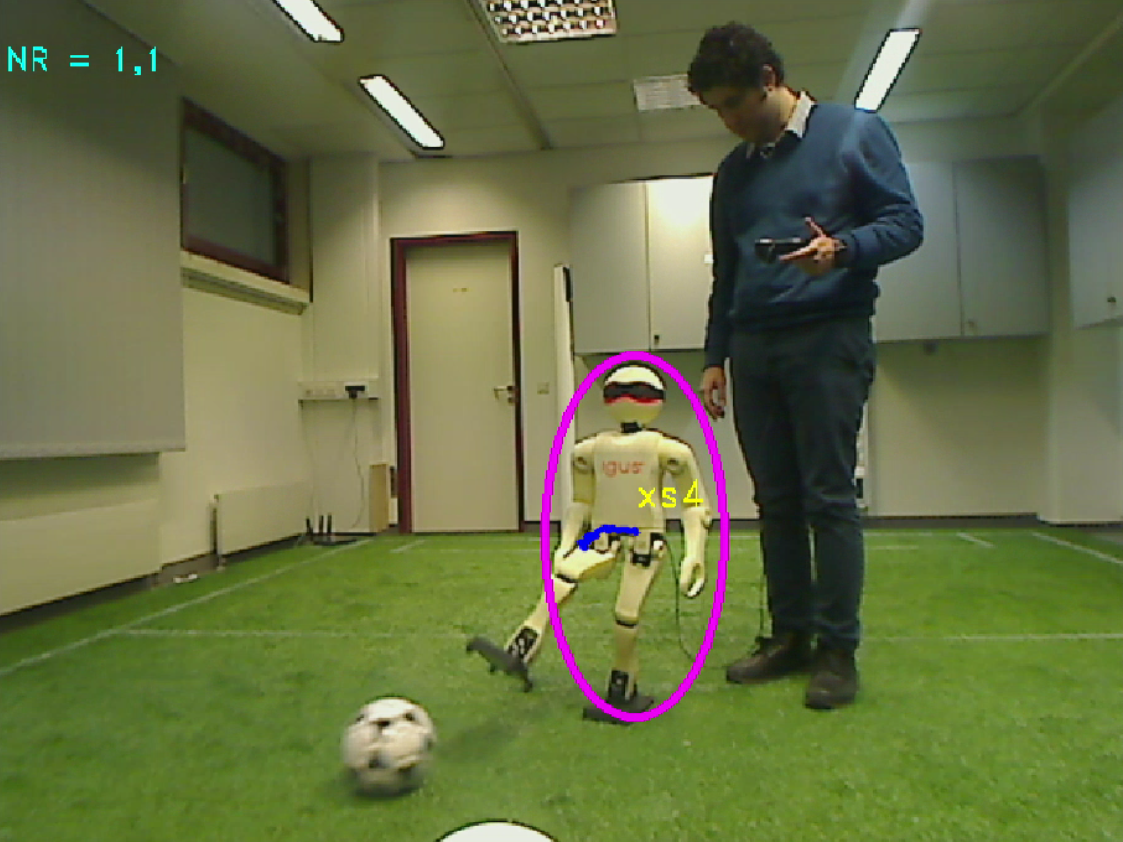}
\includegraphics[width=2.35cm]{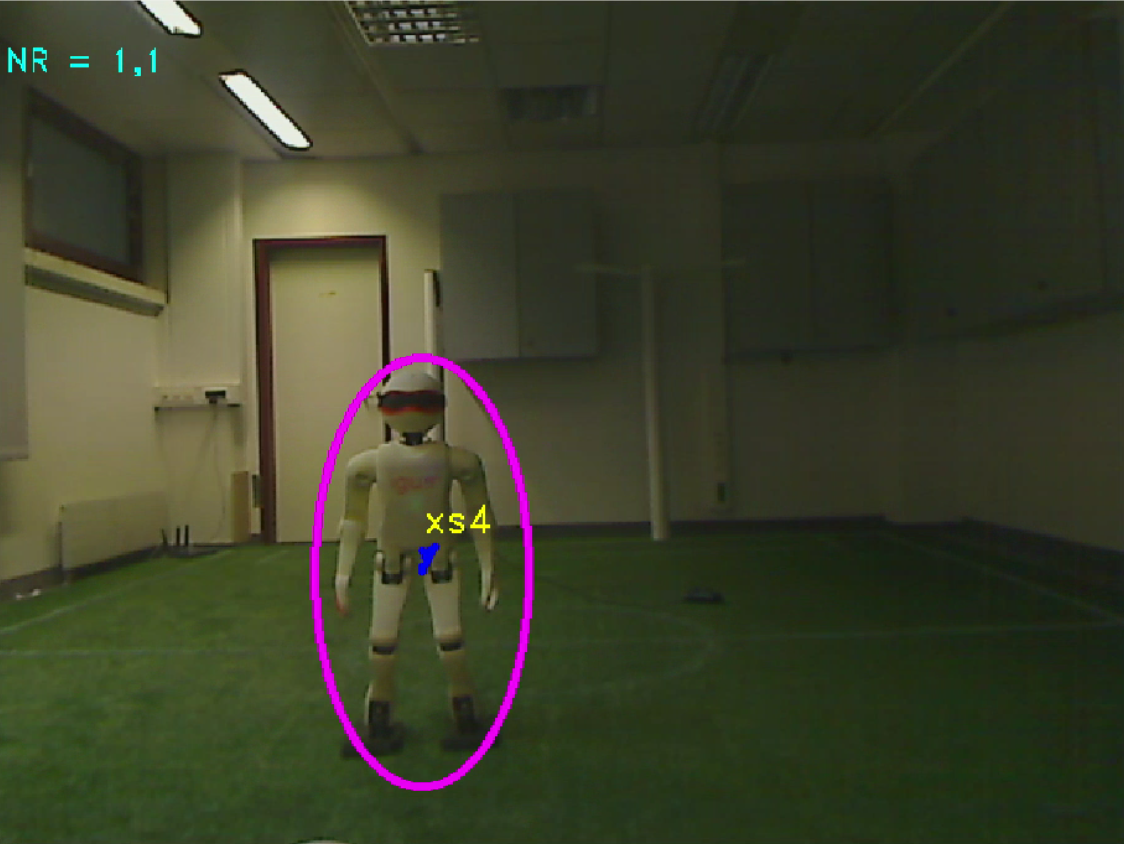}
\includegraphics[width=2.35cm]{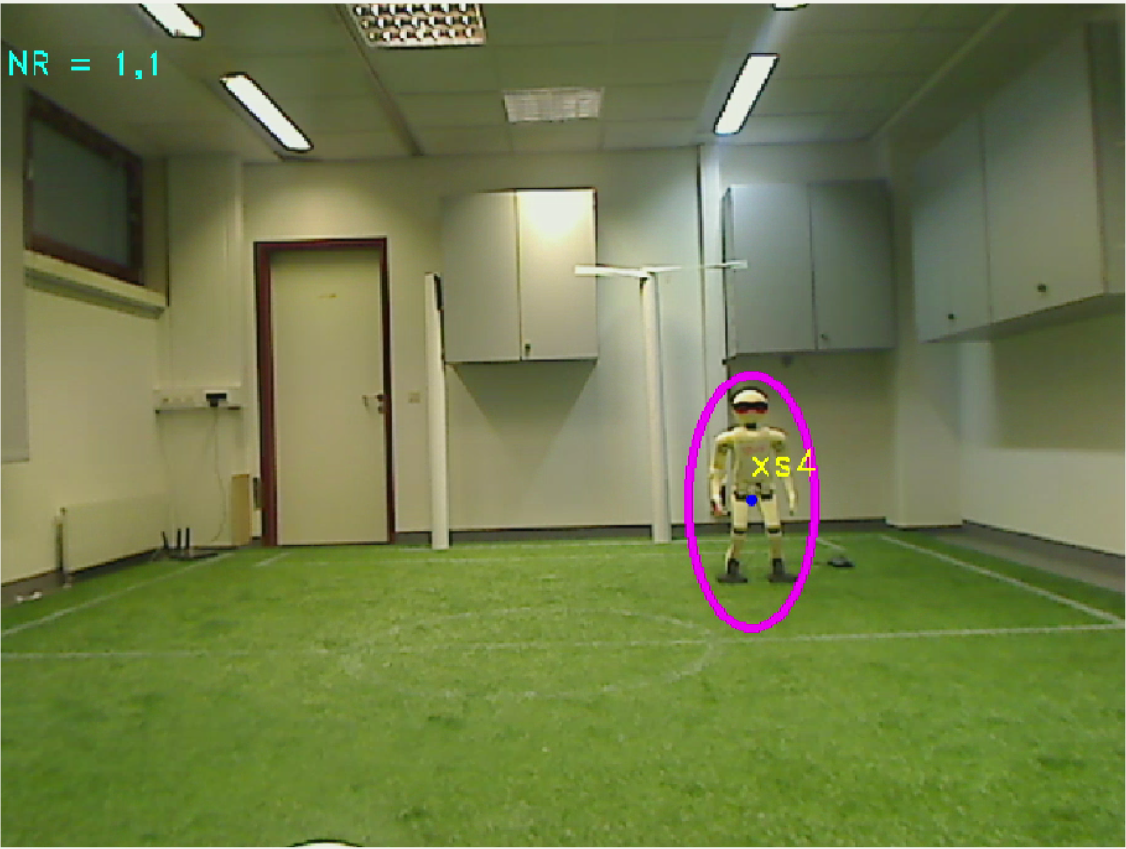}
\includegraphics[width=2.35cm]{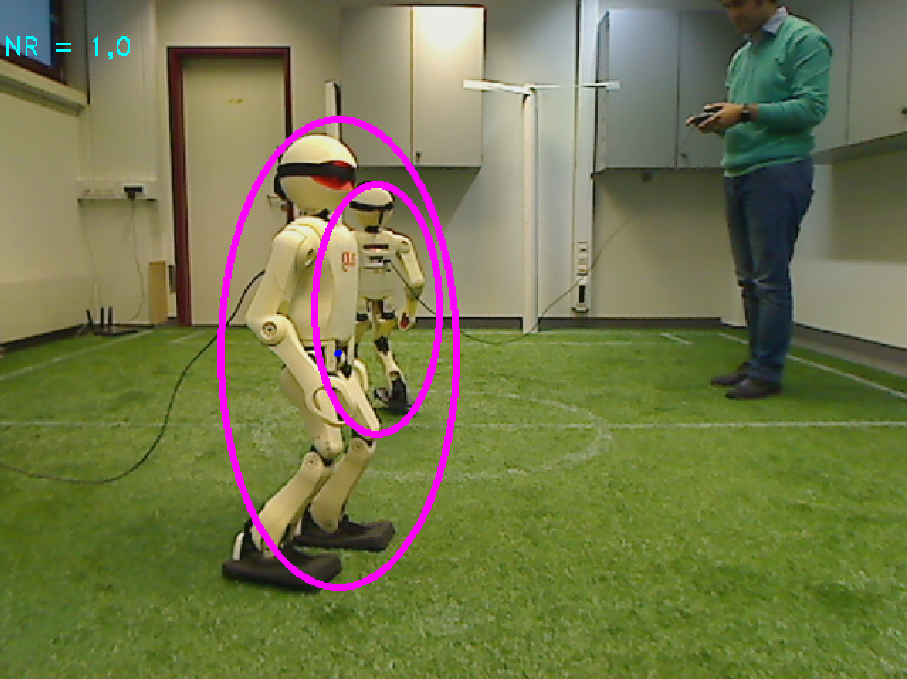}
\includegraphics[width=2.35cm]{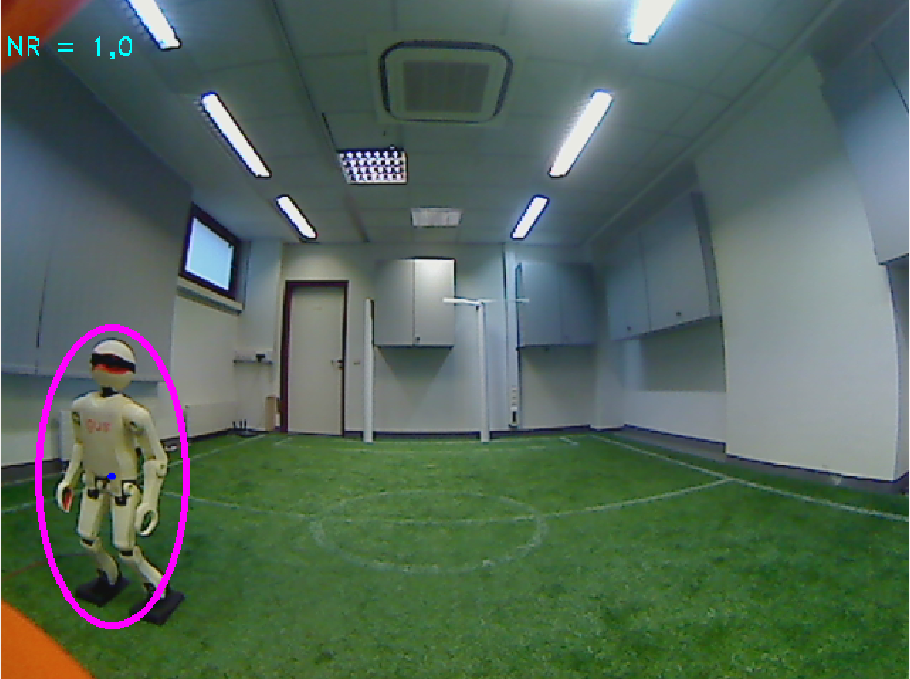}
\vspace*{-3ex}
\caption{Robot detection results under various conditions.}
\figlabel{detection}
\vspace*{-3ex}
\end{figure}

As demonstrated in \figref{detection}, this approach can detect the robot under 
various conditions, including while walking and kicking. The best detection 
results on the RoboCup field are at distances between \SI{1}{\metre} and 
\SI{5}{\metre} when the observer is not moving. After some post-processing, mainly related to non-maximum 
suppression, a bounding box for each detection is computed.

\subsection{Heading Estimation}
\seclabel{heading_estimation}

Given that all robots being detected in our application have the same visual 
appearance, estimation of the robot heading relative to the observer 
forms a primary cue to identify the robots, especially after long 
occlusions. To visually estimate the robot heading, we analyze the 
bounding boxes reported by the robot detector. We formulate the 
heading estimation problem as a multiclass classification problem by 
partitioning the full heading range into ten classes of size \SI{36}{\degree}, 
and use an SVM multiclass classifier with an RBF kernel.

The estimation is performed based on the output of a dense HOG descriptor on the 
upper half of the bounding box, the center position of the bounding box, and potential color 
features. The dense HOG features are used in the heading estimation to represent 
the visual features of each rotation class in the grayscale channel. Visual features of the robot are 
different depending on the position of the robot in the image. To address this 
issue, we pass the normalized position of the detected robot to our 
classifier. Many robots, including ours, have color features that can be used to 
help classify the robot heading. Hence, dense 
HOG features are also computed on the H channel, and the resulting feature 
vector is forwarded to the SVM classifier. To acquire the best possible results 
from the classifier, implemented using the LIBSVM library \cite{CC01a}, all 
feature data is linearly scaled to the unit interval, and k-fold 
cross-validation and grid searching was used to find the best parameter set.

\subsection{Foot Detection}
\seclabel{foot_detection}

\begin{figure}[!tb]
\centering
\includegraphics[width=2.84cm]{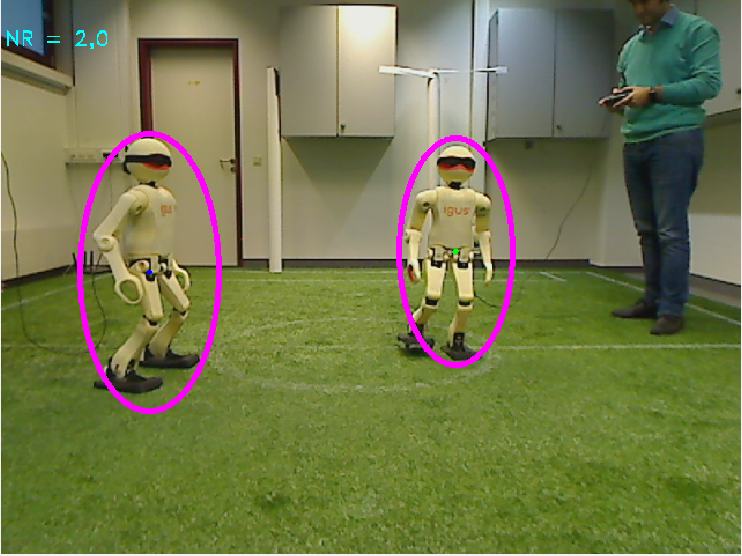}\hspace*{5ex}
\setlength{\fboxsep}{0pt}%
\setlength{\fboxrule}{0.7pt}%
\fbox{
\includegraphics[width=1.83cm]{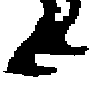}}
\fbox{\includegraphics[width=1.7cm]{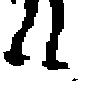}}
\vspace*{-2ex}
\caption{Two non-green binary images for the detected robots in the left image.}
\figlabel{green_binary}
\vspace*{-2ex}
\end{figure}

Once a robot has been detected, it is desirable to be able to project the 
position of the robot to the egocentric world coordinates of the observing 
robot. For this to be reliable, a good estimate of the lowest part of the detected 
robot is required. Due to the non-maximum suppression in use, the bounding box 
often may not include all pixels of the robot feet. Due to the high 
sensitivity of the projection operation, especially when the robot is far from 
the observer, this causes significant errors in the estimated  
robot distance. To overcome this problem, we make the assumption that the robot is 
located on a surface of a mostly uniform known color. Starting from an 
appropriate region of interest, and using erosion, dilation and color 
segmentation techniques, we construct a segmented binary image such as the one 
in \figref{green_binary}. A horizontal scan line scheme is then applied to 
improve the estimate of the bottom pixel of the robot. In more complicated 
cases, outside of the context of RoboCup, in which it is not possible to rely on 
a single predefined field color, one could use a background-foreground 
classification approach similar to the one proposed in \cite{milan2015joint}. 
After building a probability image, where each pixel contains the probability 
that it belongs to the background, our proposed method can be applied.

\section{Tracking and Identification System}
\seclabel{tracking_system}

Many previous works in the area of tracking and identification are not suitable 
for our application, because they either work offline or are too computationally 
expensive. In this work, we propose a real-time two-step tracking system that 
first constructs low-level tracklets through data association, and then merges 
them into tracks that are labeled with a robot ID based on tracklet angle 
differences and the reported robot heading information. For the low-level 
tracking, we use greedy initialization, albeit with the assumption that the new 
tracklet should not be in the vicinity of another existing tracklet, in which 
case lazy initialization ensures that the detection is a robot and not a false 
positive. We use lazy deletion to cope with occlusion and false negatives.

\subsection{Kalman Filter}
\seclabel{kalman_filter}

Kalman filters are a state estimation technique for linear systems, with the 
general assumption that process and observation noise are Gaussian. Many researchers utilize Kalman 
filters as part of their object tracking pipeline, mainly due to its simplicity 
and robustness. Kalman filtering involves two main steps: Prediction and 
correction. In each cycle, a new location of the target is predicted using the 
process model of the filter, and in every frame where we detect a target, we 
update the corresponding Kalman filter with the position of the detection to 
correct the prediction. Using this approach, the target can still be tracked 
even if it is not detected or occluded. We use the constant acceleration model 
to derive the predictions in our model, in the one-dimensional case: 
\begin{equation}
\begin{split}
p_{k+1} &= p_k + \dot{p}_k \Delta T + \tfrac{1}{2} \ddot{p}_k \Delta T^2,\\
\dot{p}_{k+1} &= \dot{p}_k + \ddot{p}_k \Delta T, \\
\ddot{p}_{k+1} &= \ddot{p}_k,
\end{split}
\eqnlabel{motion_model}
\end{equation}
where $p_k$, $\dot{p}_k$ and $\ddot{p}_k$ are the position, velocity and 
acceleration respectively at time step $k$. So, in our two-dimensional case the 
state vector becomes
\begin{equation}
\boldsymbol{x}_k = 
\begin{bmatrix}
h_k \,&\, v_k \,&\, \dot{h}_k \,&\, \dot{v}_k \,&\, \ddot{h}_k \,&\, \ddot{v}_k
\end{bmatrix}^T\!\!,
\eqnlabel{rotmatrowcol}
\end{equation}
where $(h_k, v_k)$ is the position of the center of the robot in the image at 
time step $k$. The system model is then given by
\begin{equation}
\boldsymbol{x}_{k+1} = \boldsymbol{\Phi} \boldsymbol{x}_k + \boldsymbol{w}_k,
\end{equation}
where $\boldsymbol{w}_k \sim \mathcal{N}(\boldsymbol{0},\boldsymbol{Q}_k)$ is zero mean Gaussian process noise with covariance $\boldsymbol{Q}_k$ and $\boldsymbol{\Phi}$ is the state transition matrix, derived 
from \eqnref{motion_model}:
\begin{equation}
\boldsymbol{\Phi} = 
\begin{bmatrix}
1 & \:0 & \Delta T & 0 & \tfrac{1}{2}\Delta T^2 & 0  \\
0 & \:1 & 0 & \Delta T & 0 & \tfrac{1}{2}\Delta T^2 \\
0 & \:0 & 1 & 0 & \Delta T & 0 \\
0 & \:0 & 0 & 1 & 0 & \Delta T \\
0 & \:0 & 0 & 0 & 1 & 0 \\
0 & \:0 & 0 & 0 & 0 & 1 \\
\end{bmatrix}\!.
\end{equation}
$\Delta T$ is the nominal time difference between two successive frames. 
In every frame where the robot is detected, the Kalman filter is updated using 
the coordinates of the center of the detected robot bounding box $\boldsymbol{z}_k = 
(\hat{h}_k, \hat{v}_k)$. The measurement model is given by
\begin{equation}
\boldsymbol{z}_k = \boldsymbol{H} \boldsymbol{x}_k + \boldsymbol{\vartheta}_k,
\end{equation}
where $\boldsymbol{\vartheta}_k \sim \mathcal{N}(\boldsymbol{0},\boldsymbol{R}_k)$ is zero mean Gaussian measurement noise with covariance $\boldsymbol{R}_k$ and $\boldsymbol{H}$ is the measurement matrix
\begin{equation}
\boldsymbol{H} = 
\begin{bmatrix}
1 & 0 & 0 & 0 & 0 & 0 \\
0 & 1 & 0 & 0 & 0 & 0 \\
\end{bmatrix}\!.
\end{equation}
Given this system model, measurement model, and some initial conditions, the 
Kalman filter can estimate the state vector $\boldsymbol{x}_k$ at each time step together with its  covariance $\boldsymbol{\Sigma}_k$.

\subsection{Data Association}
\seclabel{data_association}

In multi-target tracking, the problem of finding the optimal assignment between 
new target detections and existing tracklets, in such a way that each detection 
is assigned to at most one tracklet, is referred to as the data association 
problem. Assume that in the current frame we have $n$ existing tracklets, and 
$m$ new detections, where $m$ is not necessarily equal to $n$. Let $\boldsymbol{p}_i$ denote 
the predicted position of the $i^{th}$ tracklet, and $\boldsymbol{d}_j$ denote the position 
of the $j^{th}$ detection. We construct the $n\times m$ cost matrix $\boldsymbol{C}$, with 
entries given by
\begin{equation}
C_{ij} = 
\begin{cases}
\norm{\boldsymbol{p}_i - \boldsymbol{d}_j} & \text{if $\norm{\boldsymbol{p}_i - \boldsymbol{d}_j} < D_{max}$}, \\
C_{max} & \text{otherwise},
\end{cases}
\eqnlabel{centries}
\end{equation}
where $i = 1, \ldots, n$ and $j = 1, \dots, m$, $D_{max}$ is a distance threshold, and 
$C_{max}$ is the length of the diagonal of the image in units of pixels. Using 
the cost matrix $\boldsymbol{C}$, the optimal data association is calculated using the 
Hungarian algorithm.

\subsection{Robot Identification}
\seclabel{robot_identification}

We modeled the problem of identifying the robots as a high-level data 
association problem. In each time step, we have $n$ tracklets and $r$ robots, 
where $r$ is determined by the observer as the number of robots that are 
broadcasting their heading information over Wi-Fi. Each tracklet, in addition to 
a buffer $T_{pos}$ of $(x,y)$ pixel position values, incorporates a buffer of 
detected robot headings $T_{rot}$. Buffers $R_{rot}$ of received 
absolute headings from the robots are also maintained. The previously calculated 
robot positions are also kept in a buffer $R_{pos}$ of pixel position values. We 
wish to optimally assign each tracklet to at most one robot, based on the 
detected and received heading information \figref{identification}. The core idea is to find the best 
tracklet assignments based on the average of the differences between the 
detected tracklet heading buffers and the broadcasted headings from the 
individual robots over a limited time range. We construct the $n \times r$ cost 
matrix $\boldsymbol{G}$, with entries $G_{ij}$ that relate to the cost of associating the 
$i^{th}$ tracklet with the $j^{th}$ robot:
\begin{align}
\gamma &=
\begin{cases}
\tfrac{r}{2\pi} \min \Bigl\{ \bigl\lvert R^{a}_{rot}[1] - R^{b}_{rot}[1] \bigr\rvert : a < b, \;a, b \in 1, \ldots, r \Bigr\} & \text{if $r \geq 2$,}\\
0.5 & \text{otherwise,}
\end{cases} \\
G_{ij} &=
\begin{cases}
\frac{\gamma}{\pi D_i}\sum\limits_{k=1}^{D_i}{\bigl\lvert T^i_{rot}[k] - R^j_{rot}[k] \bigr\rvert} + \frac{1-\gamma}{C_{max}} \bigl\lVert T^i_{pos}[1] - R^j_{pos}[1] \bigr\rVert & \text{if $D_i \geq \tau$,}\\
2.0 & \text{otherwise,}
\end{cases}
\end{align}
where $\tau$ is a minimum buffer size threshold, $D_i$ is the number of elements 
in the buffers of the $i^{th}$ tracklet, and for example $T^i_{rot}[k]$ is the 
$k^{th}$ element of the $T_{rot}$ buffer for the $i^{th}$ tracklet, where $k = 
1$ corresponds to the most recently added value, and $k = D_i$ corresponds to 
the oldest value still in the buffer. Similarly, $R^j_{pos}[1]$ is the most 
recent $(x,y)$ coordinate in the $R_{pos}$ buffer for the $j^{th}$ robot. The 
interpolation factor $\gamma$ determines, based on the minimum separation of the 
broadcasted robot headings, how much we should rely on differences in heading to 
associate the robots, and how much we should rely on differences in detected 
position. Once the cost matrix $\boldsymbol{G}$ has been constructed as described, the 
Hungarian algorithm is used to find the optimal robot-to-tracklet association.
With that association, all information that is required to compute 
the egocentric world coordinates of the detected robots relative to the observer 
is available. Some low-pass filtering is performed on the final world 
coordinates to reduce the effects of noise, and produce more stable outputs.
\begin{figure}[!]
\centering
\vspace*{-2ex}
\includegraphics[width=12.1cm,height=11cm]{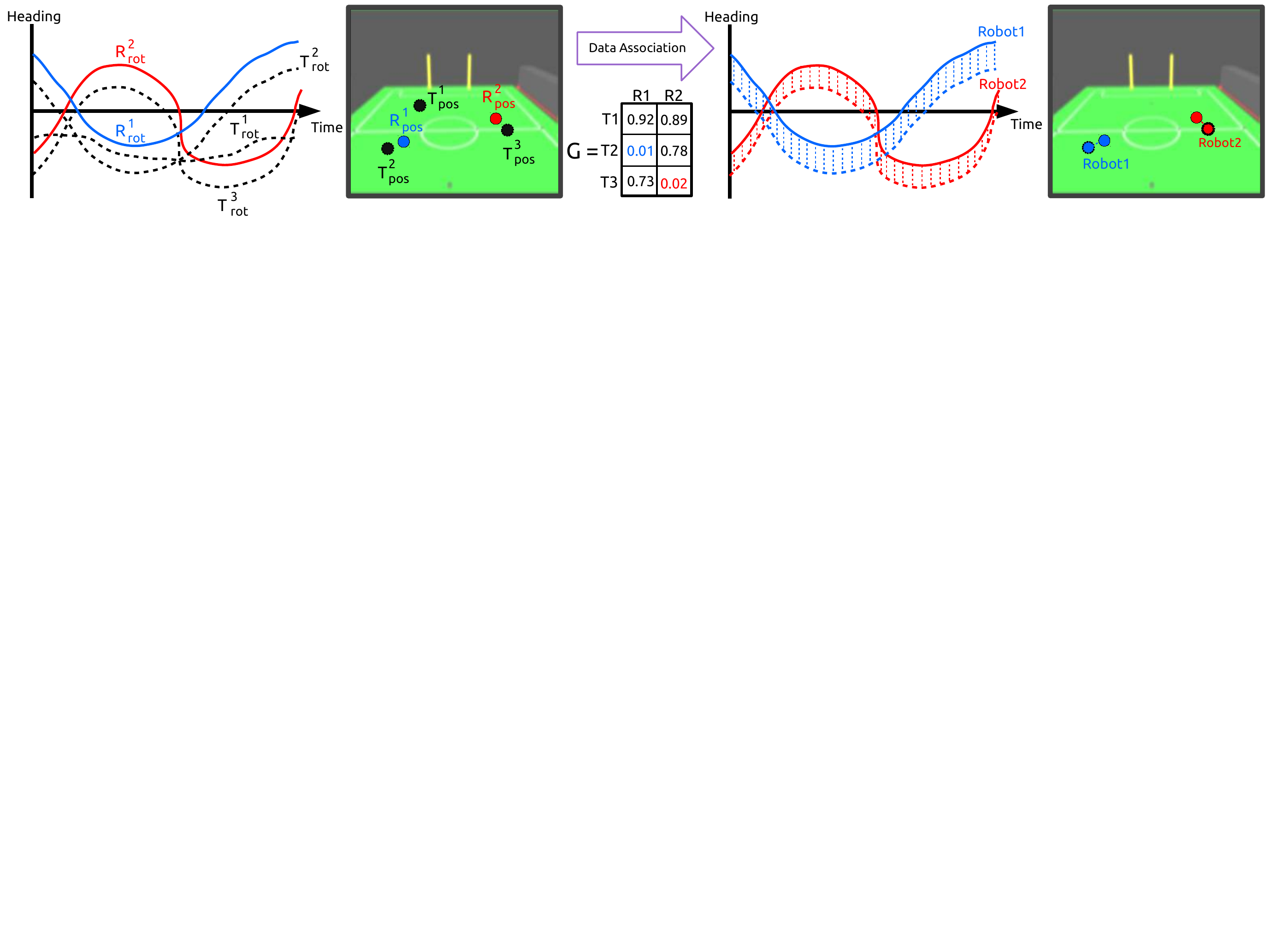}
\vspace*{-59ex}
\caption{Robot identification overview. We associate low-level tracklets with robots using comparison of heading and position.}
\vspace*{-4ex}
\figlabel{identification}
\end{figure}

\section{Experimental Results}
\seclabel{experimental_results}

In our experiments, we used two \iguhop{} robots~\cite{allgeuer2015child}. Each of 
them is equipped with a dual-core i7-4500U \SI{2.4}{GHz} processor and a 720p 
Logitech C905 USB camera. On this hardware, the whole detection, tracking and 
identification pipeline takes around \SI{50}{\milli\second}, making it suitable 
for real-time applications. We performed four different tests to evaluate the 
proposed system. All tests were conducted on a RoboCup artificial grass field, 
and the results were manually evaluated for a subset of the frames by the 
user. The data that was used in the evaluation included varying lighting 
conditions, and partial, short term and long term occlusions. In the first 
experiment, we examined the output of the robot detection module by counting the 
number of successful detections and false positives. The second experiment 
tested the success rate of the foot detection. A detected position was declared 
successful if it was within a maximum of 8 pixels from the true bottom pixel of 
the robot. The third experiment tested the success rate and average error of the 
visual heading estimation, as compared to the ground truth heading output 
broadcasted by the corresponding robot. A success was declared if the angular 
deviation was under \SI{18}{\degree}, half the size of the heading classes. In 
the final experiment, the robot identification output was verified by counting 
the proportion of frames in which the robot labels were correctly assigned. The 
results are summarized in \tabref{results}. Note that in some of the experiments, we 
used a camera attached to a laptop, and in other experiments we used a further 
\iguhop. \figref{result} shows example results of detecting, tracking, identifying, and localizing two robots on the soccer field.

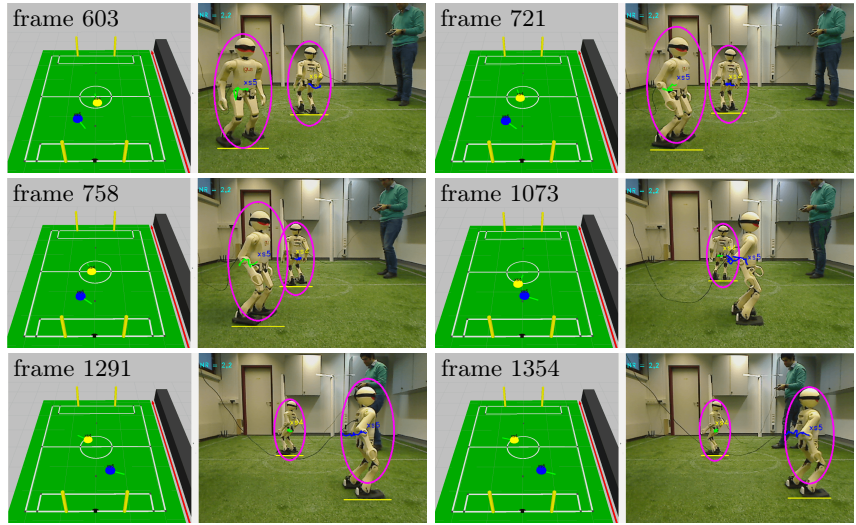
\begin{figure}[!tb]
\centering
\begin{picture}(157,65)
\put(0,0){\includegraphics[width=5.5cm]{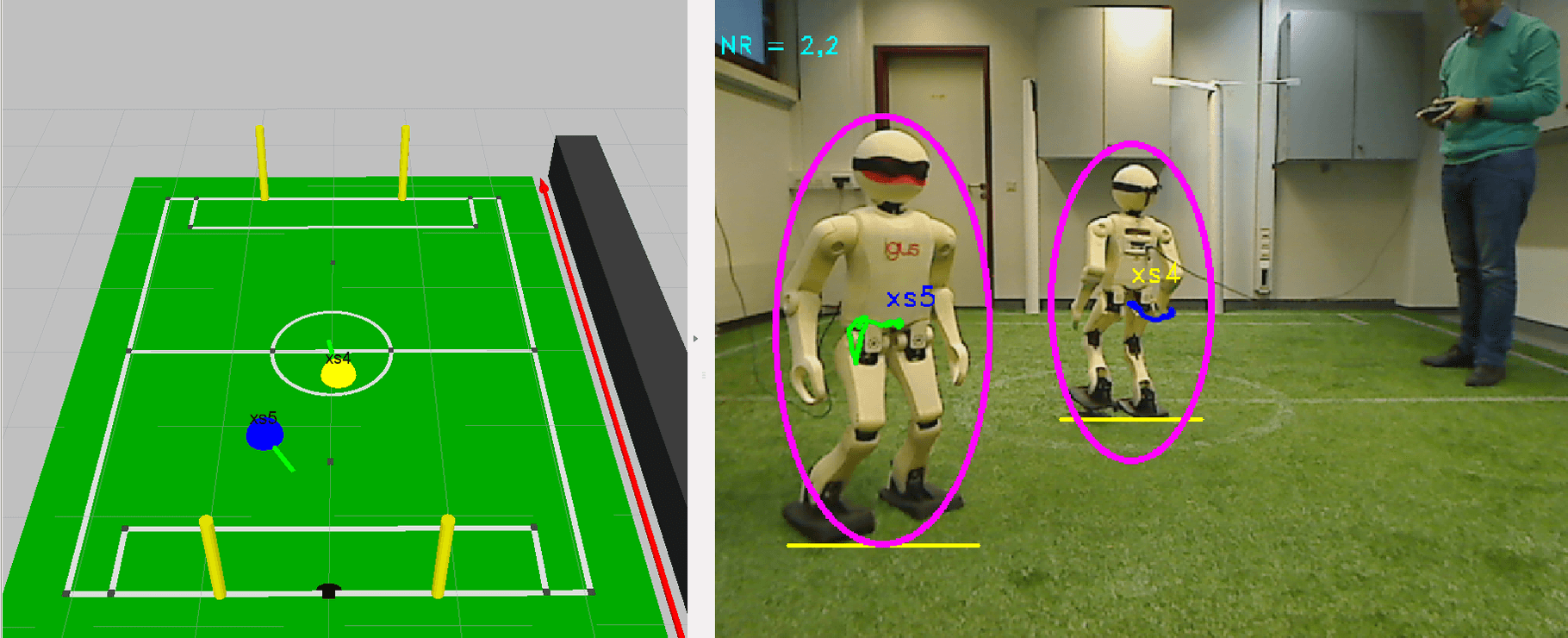}}
\put(2,55){frame 603}
\end{picture}
\begin{picture}(157,65)
\put(0,0){\includegraphics[width=5.5cm]{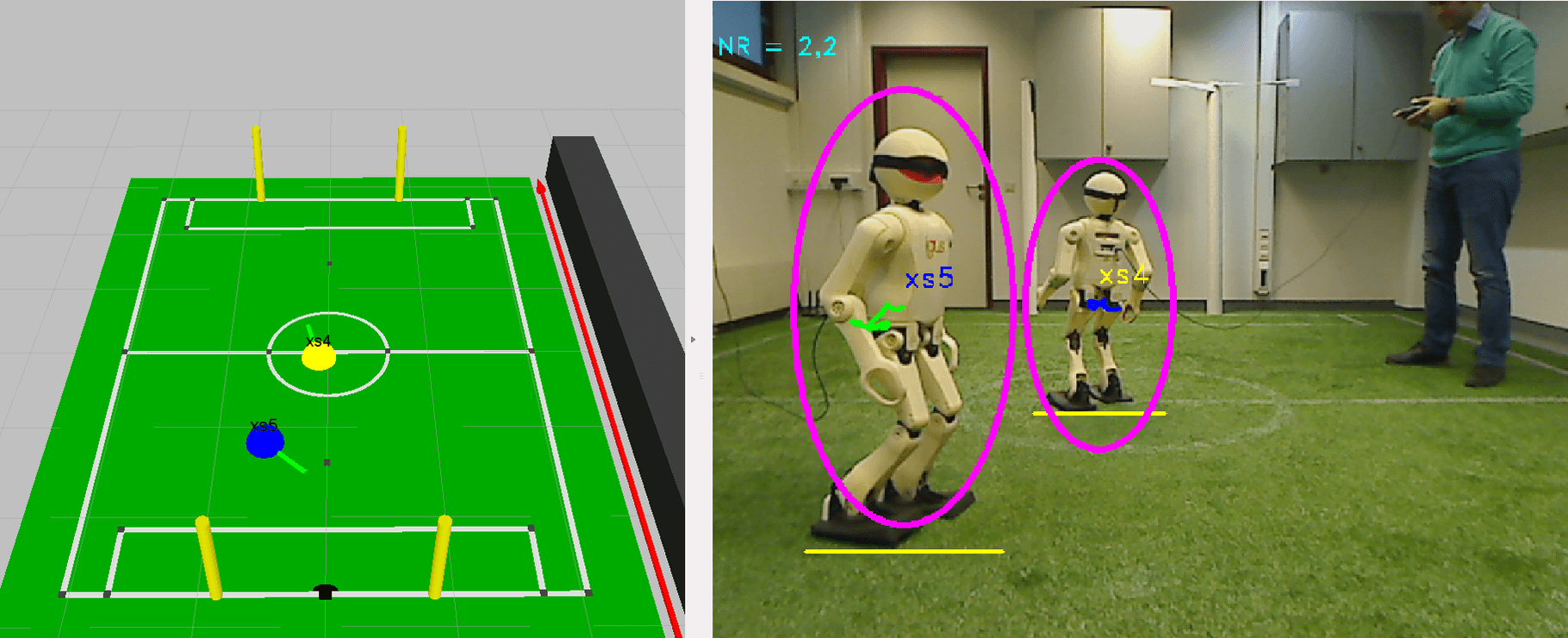}}
\put(2,55){frame 721}
\end{picture}
\begin{picture}(157,65)
\put(0,0){\includegraphics[width=5.5cm]{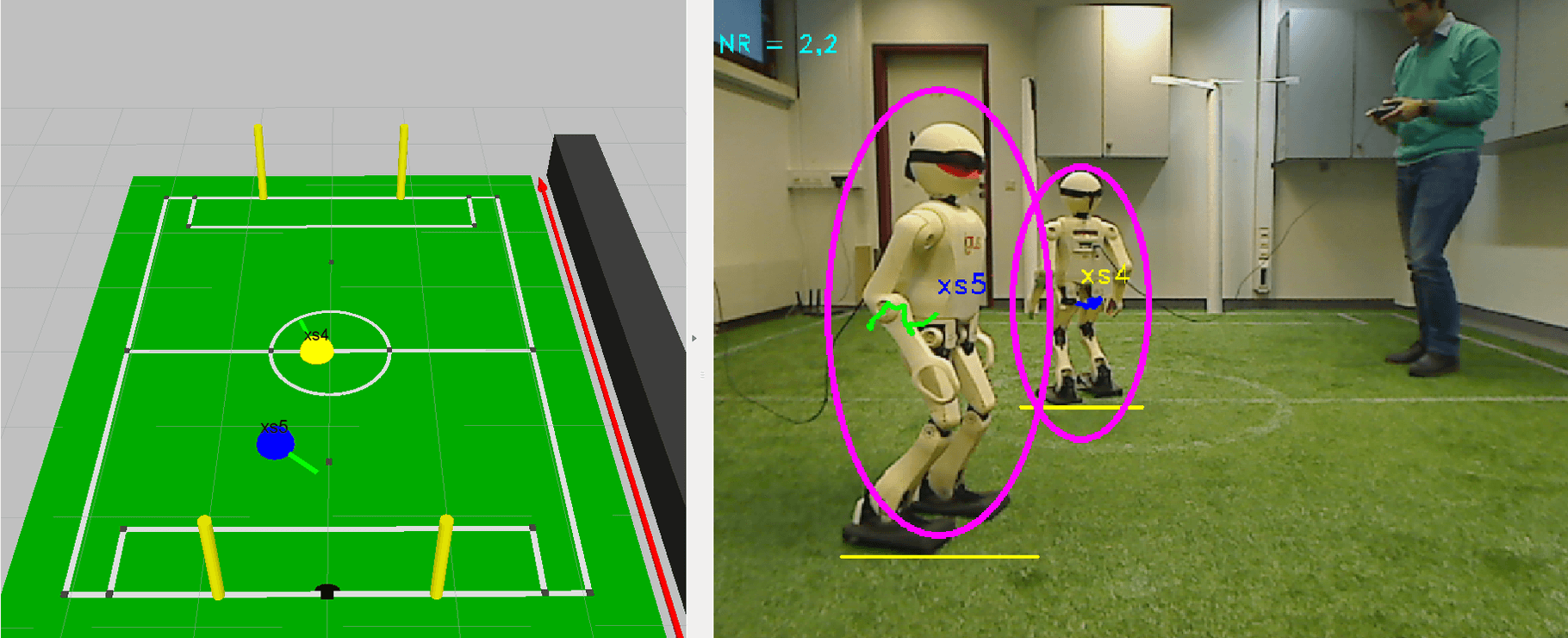}}
\put(2,55){frame 758}
\end{picture}
\begin{picture}(157,65)
\put(0,0){\includegraphics[width=5.5cm]{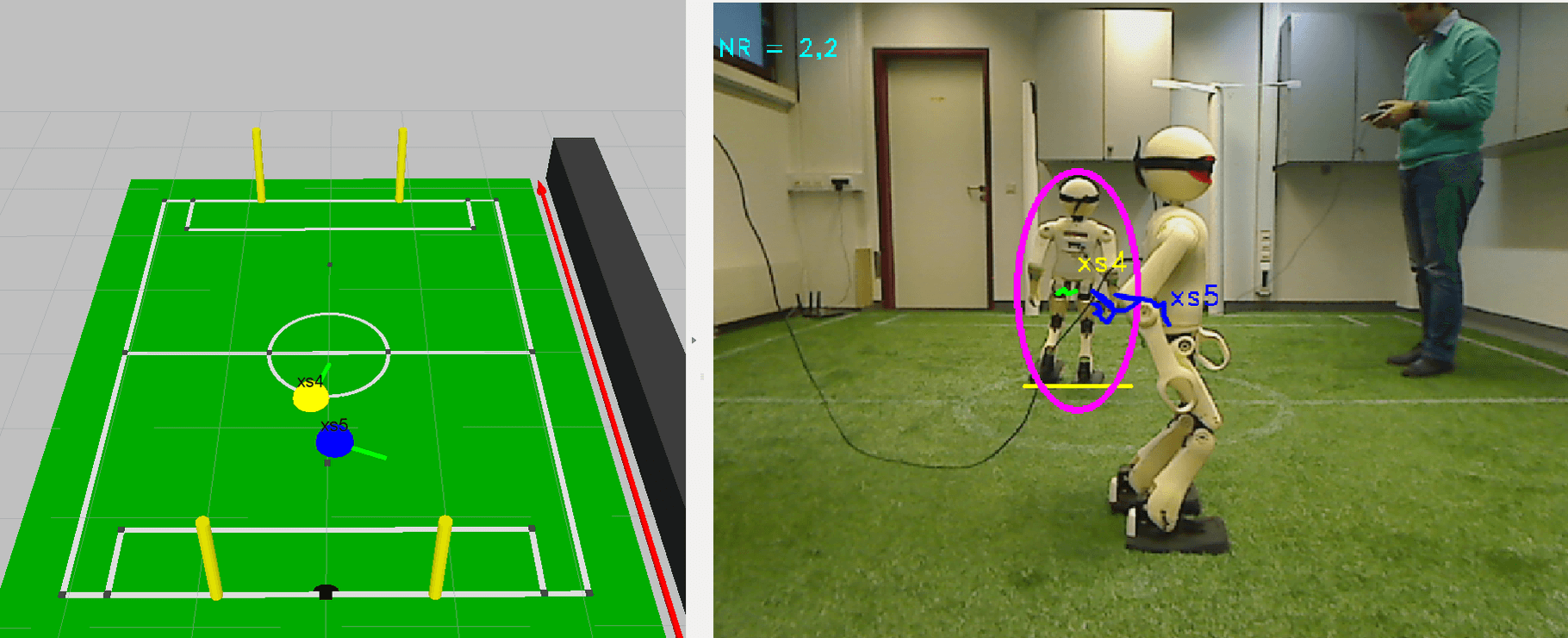}}
\put(2,55){frame 1073}
\end{picture}
\hspace*{0.5pt}
\begin{picture}(157,65)
\put(0,0){\includegraphics[width=5.5cm]{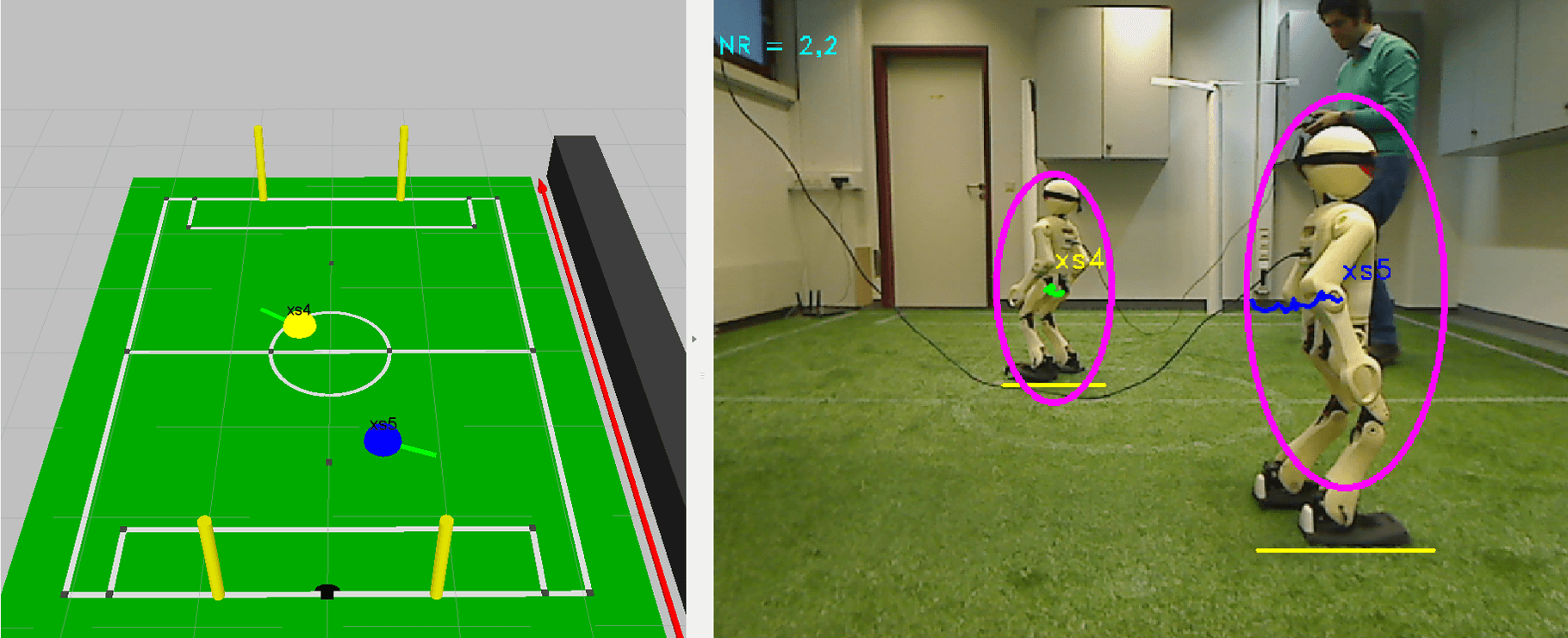}}
\put(2,55){frame 1291}
\end{picture}
\begin{picture}(157,65)
\put(0,0){\includegraphics[width=5.5cm]{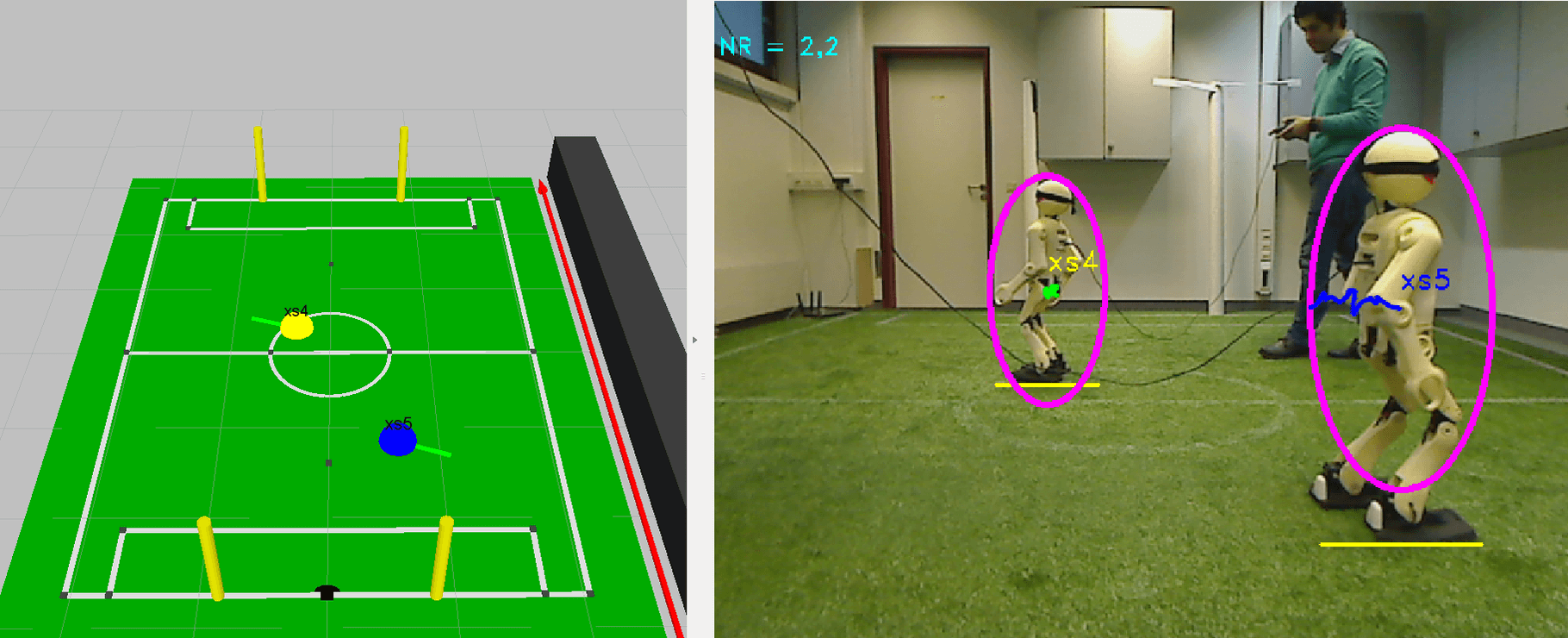}}
\put(2,55){frame 1354}
\end{picture}
\vspace*{-2ex}
\caption{Detection, tracking, and identification results obtained by our system.}
\figlabel{result}
\vspace*{-1ex}
\end{figure}

As an extension of the results, we conducted two further experiments 
where the final robot locations were broadcasted by the observer, and the robots 
used solely this localization information to walk to a predefined location on 
the field \figref{blind}. A video of the expriment is available at our website\footnote{Video link: \url{https://www.ais.uni-bonn.de/videos/RoboCup_Symposium.2016}} The cameras of the robots were covered to demonstrate 
that they were not using their own visual perception.

\begin{figure}[!]
\vspace*{-2ex}
\centering
\includegraphics[width=10cm,height=7cm]{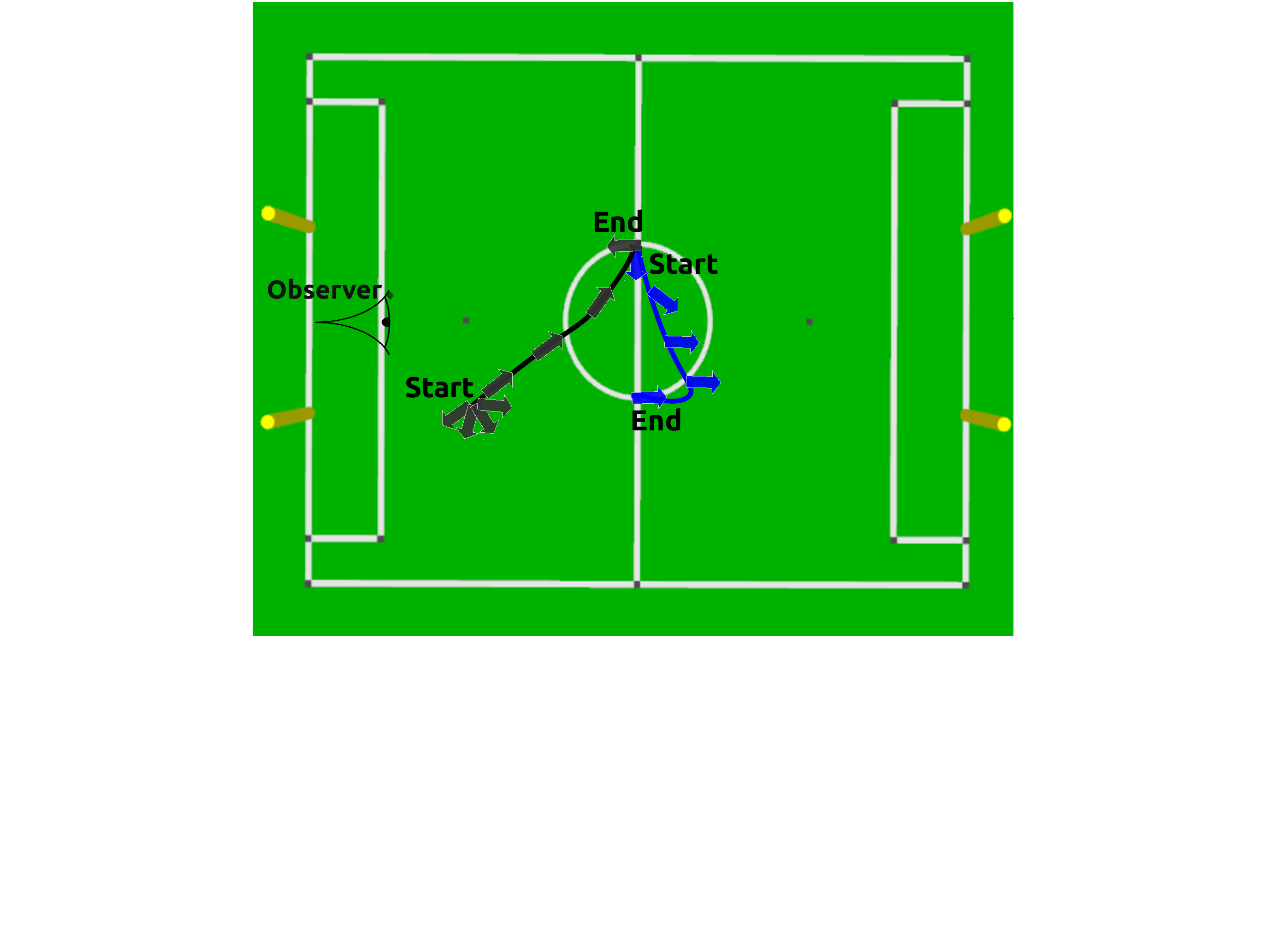}
\vspace*{-18ex}
\caption{Positioning experiment with blindfolded robots.}
\figlabel{blind}
\vspace*{-2ex}
\end{figure}

\begin{table}[!t]
\vspace*{-2ex}
\renewcommand{\arraystretch}{1.1}
\caption{Robot detection, heading estimation, and identification results.}
\tablabel{results}
\centering
\begin{tabular}{| l | c c c c |}
\hline
Test & \:Success rate\: & \multicolumn{1}{|c|}{\:False positives\:} & \multicolumn{1}{c|}{\:Average error\:} & \:Frames\:\\
\hline
Robot detection & 88\% & 7 & -- & 1000\\
Foot detection & 89\% & -- & -- & 932\\
Heading estimation & 74\% & -- & \SI{17}{\degree} & 845\\
Robot identification & 90\% & -- & -- & 932\\
\hline
\end{tabular}
\vspace*{-3ex}
\end{table}

\section{Conclusions}
\seclabel{conclusion_and_future_work}

In this paper we proposed a real-time vision pipeline for detecting, tracking, 
and identifying a set of homogeneous humanoid robots, and gained promising 
results in experimental verification thereof. Unlike many other works, we could 
not use any visual robot differences to cope with partial or complete 
occlusions, so we exploited a heading estimator to identify and track each 
robot. The result can be used in many RoboCup and real-world scenarios, such as 
for example shared localization on a soccer field, external robot control, and 
the monitoring of a group of humanoid robots using a standard camera. As future 
work, we would like to extend the robot identification to use additional data 
association cues, such as for example if a robot has fallen down or left the 
field. Additionally, we would like the observed robots to use their resulting 
tracked location to improve their own localization.

\section{Acknowledgment}
\seclabel{acknowledgment}
This work was partially funded by grant BE 2556/10 of the German Research Foundation (DFG).
The authors would like to thank Philipp Allgeuer for help in editing the article and assisting in performing experimental tests.
\bibliographystyle{abbrv}
\bibliography{document}
\end{document}